\documentclass[sigconf]{acmart}
\usepackage[capitalize,noabbrev]{cleveref} 
\begin{document}

\title{Explanova: Automatically Discover Data Insights in N × M Table via XAI Combined LLM Workflow}

\author{Yiming Huang}
\affiliation{%
  \institution{HKUST(GZ)}
  \city{Guangzhou}
  \country{China}}
\email{yhuang033@connect.hkust-gz.edu.cn}

\begin{abstract}
Automation in data analysis has been a long-time pursuit. Current agentic LLM shows a promising solution towards it. Like DeepAnalyze, DataSage, and Datawise. They are all powerful agentic frameworks for automatic fine-grained analysis and are powered by LLM-based agentic tool calling ability. However, what about powered by a preset AutoML-like workflow?  If we traverse all possible exploration, like Xn itself`s statistics, Xn1-Xn2 relationships, Xn to all other, and finally explain? Our Explanova is such an attempt: Cheaper due to a Local Small LLM.
\end{abstract}

\maketitle

\section{Introduction}
Automatic data science has been a long-time pursuit since the AutoML~\citep{automl1, automl2} era. Using an LLM-based solution raises more and more concerns due to it can convert data insights to readable human natural language. From this line, current data science agent~\citep{deepana, datawise} shows powerful analysis ability in fully automated settings, by benefiting from the strong tool calling and reasoning enhancement from RL-based post-training~\citep{rlvr}. This also results in expensive LLM exerting consumption, especially the Proprietary API ones.

Therefore, rather than improving the LLM side, we are wondering if we could enhance LLM-based automatic data science from the data analytics side. That is:
\begin{quote}
    \textit{Can preset workflow empower common LLMs' performance in automatic data analysis?}
\end{quote}

We notice that AutoML~\citep{automlzero,autosklearn} is a useful helper, since it can automatically traverse all possible relationships across data features. The rationale behind this is that statistical analysis and machine learning modeling are far cheaper than LLM calling, and remaining LLM-needed procedures can be easily handled by parallel calling. This makes data analysis automation realistic in limited GPU resource scenarios. 
\begin{figure}
    \centering
    \includegraphics[width=1.0\linewidth]{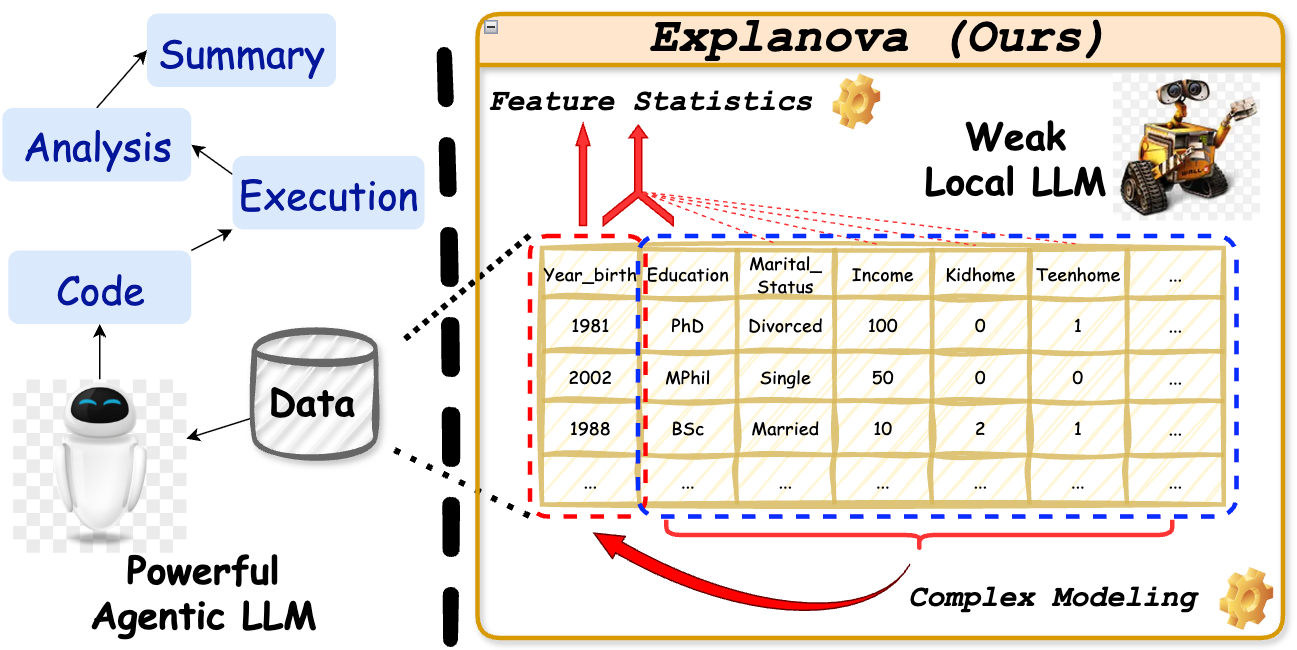}
    \caption{Illustration of our motivation.}
    \label{fig:mov}
\end{figure}

Based on that, we propose \textit{Explanova} as~\Cref{fig:mov} shows, an automatic LLM workflow that enables discovering data insights through explainable AI (XAI)~\citep{XAI} paradigm. To be specific, we mainly target single-feature statistic analysis, feature-to-feature relation statistic analysis, and feature modeling by all other features in one data table. We preset \textit{Explanova} workflow to explore all possible analytical items. 

In our workflow design, we mainly divide data automation into three stages:
\begin{enumerate}
    \item Feature Preparation Stage: We mainly focus on data pre-processing and single feature statistics in this stage.
    \item Feature-to-Feature Statistics Stage: It turns to computing statistics between any feature pair, such as correlation.
    \item Feature Modeling Stage: We use a common machine learning algorithm to predict a given feature by all remaining feature in the data table and then explain through the Shapley-value-based algorithm.
\end{enumerate}
Through the traverse in these three stages, we believe that common and valuable hidden rules in a given data table are all verified.

% Temp
The experiment shows that our proposed workflow can automatically mine the hidden rules in any given data table. Benefit from parallel design, our workflow is light and efficient to deploy on consumer-level GPU, which makes local LLM-based data analysis assistance realistic. The generated data report is detailed with a significance or credibility score for further use.

In conclusion, our contribution can be summarized in following two aspects:
\begin{itemize}
    \item Rather than the previous mindset, mainly focused on enhancing LLM's ability toward data analysis, we stress the viewpoint of improvement on the data analysis workflow for better LLM-based automatic data analysis.
    \item Our proposed \textit{Explanova} is able to automatically excavate data insights through 3 stages across 3 types of analysis. 
\end{itemize}

\section{Related Work}
\subsection{Automation in Data Science}
Automation is a lasting pursuit in data science. AutoML goes through many important evolutions. The original idea is to automatically mine the relations between data items and find the best model to realize it through a preset mechanism. Famous works as AutoML-WEKA~\citep{autoweka}, Auto-Sklearn~\citep{autosklearn}, HyperBand~\citep{hyperband}, AutoML-Zero~\citep{automlzero} shows effictive performance in this automation.

\subsection{LLM-based Data Science Automation}
Currently, using agentic LLM to accelerate automation in data science is gaining consensus in this field. There are many powerful implementations, and they can mainly be divided into two technical lines. The one is enhancing LLM's data-analysis specific ability, using RL-based post-training strategy to strengthen LLM's reasoning toward automatically planning the analysis steps, like the work DeepAnalyze~\citep{deepana}. The other one concentrates on the design of the single-agent or multi-agent framework; they mainly devote efforts to make agentic power work more reasonably, like the work Datawise Agent~\citep{datawise}. 

Besides the technical method, there are also many useful benchmarks for comparison. They are mainly organized in questions, accompanied by a data file style. For the insight level benchmark, DiscoveryBench~\citep{discoverybench} and InfiAgent-DABench~\citep{DABench} are suitable.

\subsection{XAI \& Data Science}
Explainable AI (XAI) has been involved in a deeper reveal in feature modeling. Representative works like LIME~\citep{LIME}, SHAP~\citep{SHAP1,SHAP2} show powerful explanatory ability on a given classifier or regressor, especially in model-agnostic cases.
We select Shapley-value-based SHAP due to its solid game theory basis. Meanwhile, there are mature implementations among these Shapley-based methods, the SHAP library~\citep{lundberg2017unified}. Specifically, we choose the kernel shape in our framework.

\section{Workflow Design}
\begin{figure*}
    \centering
    \includegraphics[width=1.0\linewidth]{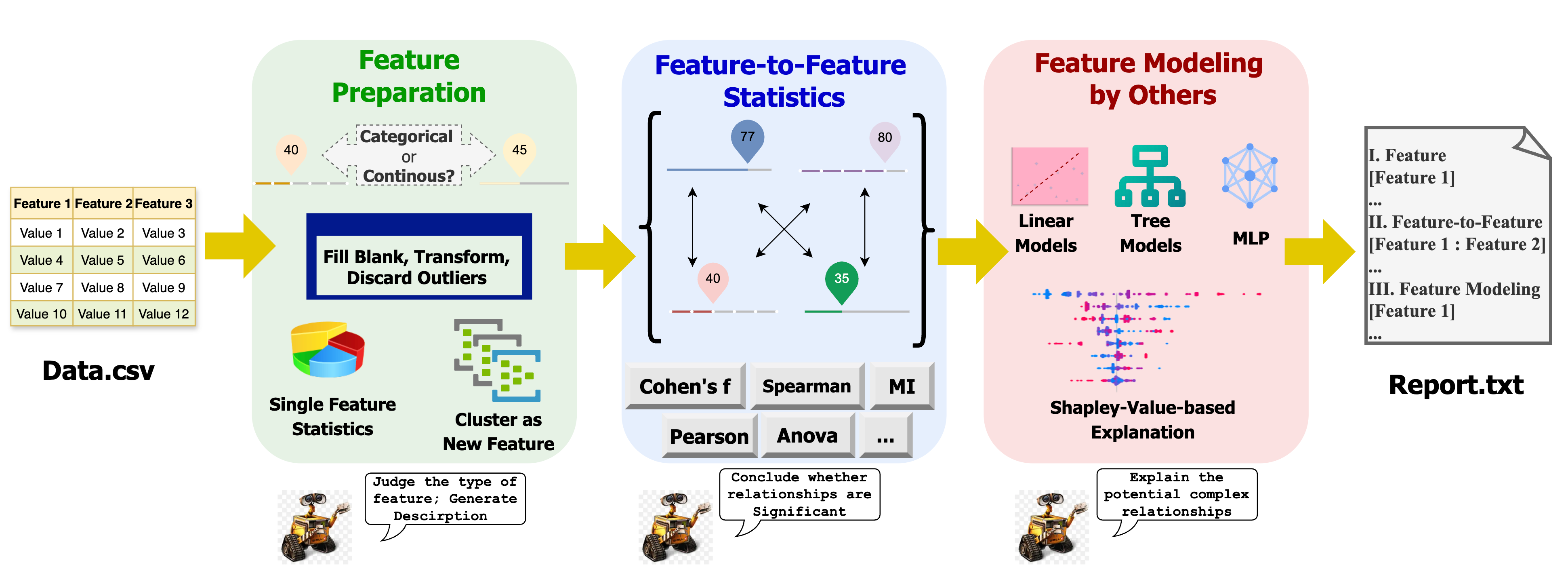}
    \caption{Demostration of our designed workflow.}
    \label{fig:arch}
\end{figure*}
\subsection{Overview}
As shown in~\Cref {fig:arch}, our Explanova is mainly performed through three stages: Feature Preparation, Feature-to-Feature Statistics, and Feature Modeling. Firstly, in the Feature Preparation stage, the feature type will be judged by LLM, and missing values, anomaly values will be handled. In addition, we perform clustering as a new feature added to this data table. Next stage, we mainly focus on exploring feature-to-feature relationships, and statistical factors are computed to judge potential statistical level relationships according to different data types. As for the feature modeling stage, our main goal is to reconstruct one feature by all remaining features so that we can get complex internal working between features. Then, we use Shapley-based explanation to extract such complex relationships. Due to error accumulation in this stage, we comprehensively use a credibility score to tell LLM the accountability of each finding. Finally, all findings are summarized in a report by LLM-converted natural language form, which shows good readability as an explanation to users.
\subsection{Feature Preparation}
\paragraph{Data Type Detection \& Transformation} Give a N $\times$ M table, we use an LLM to infer feature types from column names, first-row values. According to different pre-possess strategies, we classified them into these target types: continuous numeric, discrete numeric, non-numeric categorical, ID-like, and time-related (year or day-month-year, etc.). We drop the meaningless ID-like feature since it doesn't work for insight mining. All features will finally be transformed to only two types: Continuous \& Discrete. All LLM calls run in parallel since columns are independent in this step.

\paragraph{Missing Value Handling} To missing values in given feature, if the missing ratio is bigger than our preset threshold, we will drop this feature as this feature is regard no contribution for any modeling and analysis. Otherwise, we impute the mean or median for missing value replacement.

\paragraph{Outlier Handling} Likewise, if the missing value, If the outlier ratio exceeds the threshold we will delete through flag at these outliers.

\paragraph{Statistical Summary} We get a single feature statistic in this step. If a feature is a continuous type, we will compute its maximum/minimum value, range, variance, mean, quartiles \& IQR information. Otherwise, if it is discrete, we will get its frequency \& relative proportions, number of categories in the data.

\paragraph{Cluster-Based New Feature} To get a unified cluster feature across continuous and discrete features, we compute a Gower distance matrix (mixed-type compatibility) to run HDBSCAN~\citep{hdbscan} on the distance matrix. If the clustering error (SSE error) exceeds the threshold, we will discard this cluster feature.

Finally, we let the LLM regenerate a feature description for final documentation.
\subsection{Feature-to-Feature Statistics}

Regarding the feature-to-feature relationships, there are four possible Variable-Type combinations:
\begin{itemize}
    \item Continuous $\rightarrow$ Continuous  
    \item Continuous $\rightarrow$ Discrete
    \item Discrete $\rightarrow$ Continuous
    \item Discrete $\rightarrow$ Discrete
\end{itemize}
All analyses can run in parallel through multiprocessing.

\paragraph{Continuous $\rightarrow$ Continuous} Three statistical relationships are computed:
\begin{itemize}
    \item Pearson Correlation: It measures linear direction \& strength (in the range of -1 to 1), which is key for detecting linear dependency.
    \item Spearman Correlation: It captures monotonic trends, meanwhile, robust to outliers. It is useful when relationships are nonlinear but monotonic.
    \item Mutual Information: It detects any nonlinear dependency. Most general-purpose measure for continuous variables.
\end{itemize}

\paragraph{Continuous $\rightarrow$ Discrete (Multiclass)} Three statistical relationships are computed:
\begin{itemize}
    \item ANOVA F-statistic: It tests if class groups have different means.
    \item $\eta^2$ (Effect Size): Fraction of variance explained by class membership.
    \item Kruskal–Wallis H: It is a non-parametric alternative and robust for non-normal, skewed, or unequal variance data.
\end{itemize}

\paragraph{Discrete $\rightarrow$ Continuous} Three statistical relationships are computed:
\begin{itemize}
    \item ANOVA F-statistic: To test mean differences across discrete groups.
    \item $\eta^2$ (Effect Size): Strength of group influence on the continuous target.
    \item Cohen’s f: Effect size for multi-group comparisons; more interpretable than $\eta^2$.
\end{itemize}

\paragraph{Discrete $\rightarrow$ Discrete (Multiclass to Multiclass)} Three statistical relationships are computed:
\begin{itemize}
    \item $\chi^2$ Test of Independence: This test determines whether the two categorical variables are independent.
    \item Cramer’s V: it measures association strength (0-1), complements $\chi^2$.
    \item Mutual Information: Captures arbitrary categorical dependencies.
\end{itemize}

\paragraph{LLM-Based Statistical Relationship Description (Parallelized)} After all relation statistics are computed, for each feature, we feed all computed metrics into an LLM and automatically decide whether a significant relationship exists by preset thresholds. If significant, LLM generates a natural language description of the relationship (e.g., “Feature A increases monotonically with Feature B.”). All feature-to-feature interpretations run concurrently for scalability.

\subsection{Feature Modeling}
\paragraph{Data Rescaling} To fit the input and make sure our machine learning algorithm works properly, we execute the data rescaling as the input feature to ML models. For the continuous variables, we perform max–min scaling to [0, 1]. To categorical variables (n classes) we convert them to one-hot vectors of length n. Then, we concatenate scaled continuous and one-hot discrete features into one feature vector for prediction. As for the expected outputs: if the target feature is discrete, we perform the classification modeling, that is, the final one-hot labels. On the other hand, it will be the regression modeling that yields scalar targets.

\paragraph{Model Selection \& Parallel Training} We train 4 common model families $\times$ n features in parallel (multiprocessing). For Discrete Targets (Multiclass), these models are Multiclass Logistic Regression, Decision Tree, MLP, ensemble of the above models. For Continuous Targets, these models are Linear Regression, Decision Tree Regressor, MLP Regressor and an ensemble of the above. We use a 5-fold CV for all models and evaluate with the NLL metric. We select the best model per feature and store its uniform NLL score for credibility score computation.

\paragraph{NLL as the unified error metric} The output distributions of the abovemention models are: (Classification) model outputs class probabilities $p(y=c\mid x)$ \& (Regression) model outputs mean and variance $\mu(x),\ \sigma^2(x)$. We treat them as probability models: (Classification) $P(y_i\mid x_i)=p_\theta(y_i\mid x_i)$ (Regression, Gaussian)
$$
  P(y_i\mid x_i)=\frac{1}{\sqrt{2\pi\sigma_i^2}}
  \exp\!\left(-\frac{(y_i-\mu_i)^2}{2\sigma_i^2}\right)\notag
$$
Therefore, we can gain the sample-level NLL as metric: (Classification) $\mathrm{NLL}_i=-\log p_\theta(y_i\mid x_i)$
(Regression)
$$
  \mathrm{NLL}_i=\frac{1}{2}\log(2\pi\sigma_i^2)
  +\frac{(y_i-\mu_i)^2}{2\sigma_i^2}\notag
$$
Based on that, the overall NLL is:
$$
\mathrm{NLL}
=\frac{1}{N}\sum_{i=1}^N \mathrm{NLL}_i\notag
$$
This provides a unified interpretation that the smaller NLL the better model, allowing consistent comparison across classification and regression tasks.

\paragraph{SHAP Explanation \& Stability Evaluation} Explanova parallelize all n features in SHAP runs $\times$ k folds. We first perform basic SHAP Fit Use KernelSHAP~\citep{kernelshap} on full data points in data table. We compute basic SHAP values and global SHAP contribution via mean over samples. The we use k-Fold Perturbation SHAP (K = 3) to verifiy the faithfulness of basic shap explanation. For each fold, we drop 1/k of the data and refit a SHAP explainer. Then we compute SHAP per fold and fold-level global SHAP. We define SHAP perturbed error:
$$
\text{SHAP-Error}=\frac{1}{k}\sum_{i=1}^k 
\left|SHAP^{(i)}-SHAP^{full}\right|
$$
On the other hand, explainability also consider readability beyond faithfulness, we treat readability as how outstanding feature SHAP contribution appears among all input features. This outstandingness can be measured by disorderness, thus, we define Global SHAP Entropy to represent readability. It normalizes absolute global SHAP:
$$
p_i=\frac{|SHAP_i|}{\sum_j |SHAP_j|}
$$
next, computed as:
$$
H=-\sum_i p_i\log p_i
$$
High H indicates concentrated importance pattern which holds stronger interpretability. We save global SHAP for feature-level interpretability and store full SHAP matrix for downstream local interpretability.

\paragraph{Final Credibility Score}
We consider a single findings's crediblity in feature modeling as the comprehensive error accumulation in ML modeling, shapley computing stability and readability steps. Thus, it can be measured by:
$$
\text{Score}(R)=\frac{H}{|NLL|\times|\text{SHAP-Error}|}
$$
Where H is the SHAP entropy (global interpretability strength), NLL is the absolute model likelihood loss and SHAP-Error is the k-fold SHAP perturbation error. Lower NLL, lower SHAP-Error and higher entropy means more reliable model thus more accountable final findings toward complex modeling.

\subsection{Report Generation \& QA Adaption}
In the end, we summarize all generated desription in above stages as a final report that reveals potential useful data insight in give data table. Considering the QA need in some scaniros, we design a cache mechanism that Explanova will not re-run the automatic exploration when it has the cached report w.r.t. a same name data table. It will answer question based on cached report. If a report's length exceeds LLM's context window, it will be divide into chunks to ask parallelized LLM to first gain sparate answer in individual chunks then give the final answer.

\section{Experiment}
\subsection{Implementaion Details}
We mainly adopt \verb|LangChain|, \verb|sklearn|, and \verb|shap| to implement our workflow. \textbf{all code, used prompts, and results are included in the source code zip submitted to the course project.}
\subsection{Case Study on DSAA 5002 Default Target: \textit{market\_compagian.csv}}
We apply our explanova to the \textit{market\_compagian.csv}, and the base LLM is Qwen3-8B It takes 5 minutes to generate the below result. From this result, we can see Explanova can generate a detailed report for any given data table.
{\tiny
\begin{verbatim}
================================================================================
EXPLANOVA ANALYSIS REPORT                            
================================================================================

Generated: 2025-12-06 18:18:44
Report saved to: ../output/explanova_report.txt

Dataset: This is a marketing campaign dataset containing customer information and purchase behavior
Number of features analyzed: 28
Data shape: (2240, 28)

================================================================================
5.1 INDIVIDUAL FEATURE STATISTICS
================================================================================

[Feature: Year_Birth]
  Type: continuous
  Description: The Year_Birth feature represents customers' birth years, with a mean of 75.81 and median of 77, 
  ranging from 0 to 103, indicating a continuous distribution of ages in the marketing dataset.
  Statistics:
    - Mean: 75.8058
    - Median: 77.0000
    - Std: 11.9841
    - Range: [0.0000, 103.0000]
    - IQR: [66.0000, 84.0000]
    - Variance: 143.6179

[Feature: Education]
  Type: categorical
  Description: The "Education" feature is a categorical variable with 5 categories, 
  representing customers' educational levels, with sample values indicating most customers have a 2-level 
  education and some have a 4-level education.
  Statistics:
    - Number of categories: 5
    - Category frequency:
      2: 1127 (50.31%)
      4: 486 (21.70%)
      3: 370 (16.52%)
      0: 203 (9.06%)
      1: 54 (2.41%)
    - Category mapping: {'2n Cycle': 0, 'Basic': 1, 'Graduation': 2, 'Master': 3, 'PhD': 4}
......
================================================================================
5.2 FEATURE-TO-FEATURE RELATIONSHIPS
================================================================================

Relationship Summary:
  Total relationships analyzed: 720
  Significant (above threshold): 176
  Non-significant (below threshold): 544

Significance Thresholds Used:
  Continuous-Continuous:
    - |Pearson r| >= 0.3
    - |Spearman r| >= 0.3
    - Mutual Info >= 0.1
  Continuous-Discrete / Discrete-Continuous:
    - ANOVA p < 0.05
    - eta^2 >= 0.06
  Discrete-Discrete:
    - Chi^2 p < 0.05
    - Cramer's V >= 0.1

--------------------------------------------------------------------------------
SIGNIFICANT RELATIONSHIPS (n=176)
--------------------------------------------------------------------------------

1. Year_Birth -> Education [SIGNIFICANT]
   Description: There is a significant relationship between Year_Birth and Education, 
   indicating variations in birth years across different educational levels.
   Statistics: anova_f=21.6775, anova_p=0.0000, eta_squared=0.0373, kruskal_h=88.5519

2. Year_Birth -> Marital_Status [SIGNIFICANT]
   Description: Marital status significantly influences year of birth, with notable differences 
   in birth years across marital categories.
   Statistics: anova_f=14.9060, anova_p=0.0000, eta_squared=0.0447, kruskal_h=99.6889
......
================================================================================
5.3 COMPLEX RELATIONSHIP CREDIBILITY SCORES
================================================================================

Credibility Score Formula: Score = H / (|NLL| * |SHAP-Error|)
  H: SHAP Entropy (higher = more concentrated importance)
  NLL: Model training loss (lower = better fit)
  SHAP-Error: K-fold stability (lower = more stable)

Credibility Level Thresholds:
  HIGH:   Score >= 10.0
  MEDIUM: 3.0 <= Score < 10.0
  LOW:    Score < 3.0

Features by Credibility Level:
  HIGH:   9 features
  MEDIUM: 3 features
  LOW:    7 features

All Features Ranked by Credibility Score:

Rank   Feature                   Score        Level      H          NLL        SHAP-Err    
------ ------------------------- ------------ ---------- ---------- ---------- ------------
1      Marital_Status            6798.7439    HIGH       4.0853     1.4235     0.0004      
2      Education                 2858.2658    HIGH       3.9117     1.2025     0.0011      
3      Kidhome                   1599.4103    HIGH       2.4815     0.3927     0.0040      
4      Teenhome                  1122.7494    HIGH       2.5300     0.4359     0.0052      
5      NumDealsPurchases         89.0858      HIGH       2.5358     1.6799     0.0169      
6      NumCatalogPurchases       81.9291      HIGH       2.2158     1.9348     0.0140      
7      NumStorePurchases         68.6566      HIGH       2.4030     2.0662     0.0169      
8      NumWebVisitsMonth         60.9354      HIGH       2.7083     1.7713     0.0251      
9      NumWebPurchases           57.6450      HIGH       2.2728     1.9437     0.0203      
10     Recency                   5.7753       MEDIUM     2.7522     4.7926     0.0994      
11     Year_Birth                3.7186       MEDIUM     2.5804     3.7782     0.1837      
12     MntSweetProducts          3.2248       MEDIUM     2.5848     4.8099     0.1666      
13     MntFruits                 2.2749       LOW        2.3399     4.7695     0.2157      
14     MntGoldProds              1.7045       LOW        2.6747     5.1670     0.3037      
15     MntFishProducts           1.6609       LOW        2.5433     5.0519     0.3031      
16     MntMeatProducts           0.3018       LOW        2.5272     6.2976     1.3295      
17     Dt_Customer               0.1769       LOW        2.6662     6.5979     2.2846      
18     MntWines                  0.1617       LOW        2.4007     6.6258     2.2412      
19     Income                    0.0013       LOW        2.4319     11.1625    164.5914    


Detailed SHAP Analysis for Each Feature:
--------------------------------------------------------------------------------

[1] Marital_Status (Credibility: HIGH, Score: 6798.7439)
  SHAP Entropy: 4.0853
  K-Fold Stability Error: 0.0004
  Model: logistic (NLL: 1.4235)
  All predictor features ranked by SHAP importance:
  LLM Interpretation:
    The top SHAP features for predicting Marital_Status are Income, Year_Birth, Kidhome, Teenhome, 
    and Dt_Customer, indicating their significant influence on the classification. No interval-specific 
    patterns were identified as interval analysis is disabled. The model shows high interpretability 
    with low SHAP fitting error and moderate entropy, suggesting reliable and stable feature importance 
    across different data subsets.


[2] Education (Credibility: HIGH, Score: 2858.2658)
  SHAP Entropy: 3.9117
  K-Fold Stability Error: 0.0011
  Model: logistic (NLL: 1.2025)
  All predictor features ranked by SHAP importance:
  LLM Interpretation:
    The top features driving Education predictions are Year_Birth, Income, Kidhome, Teenhome, 
    and Dt_Customer, indicating strong influence on educational outcomes. 
    No interval-specific patterns were identified as interval analysis is disabled. 
    The model shows good interpretability with low SHAP fitting error (0.0011) and 
    moderate entropy (3.9117), suggesting consistent and reliable feature contributions across the dataset.

\end{verbatim}}
%\subsection{Benchmarking in}

\section{Conclusion \& Limitiations}
We propose Explanova as another way toward automatic LLM-based data analysis automation. The experiment shows its effectiveness in generating data insights in a given data table.

Due to time limitations, there are many imperfections in this work. Such as its function needs to fix some parallel bugs in Shapley-value-based analysis by featuring different intervals and finishing the figure plotting in the final report. We also need to benchmark our Explanova: QA-form Explanova according to the final report / Comparison on InfiAgent-DABench /Comparison on DiscoveryBench.
\bibliographystyle{ACM-Reference-Format}
\bibliography{main}

\end{document}